\documentclass{article}
\usepackage[preprint]{neurips_2025}
\usepackage[utf8]{inputenc} 
\usepackage[T1]{fontenc}   
\usepackage{hyperref}       
\usepackage{url}            
\usepackage{booktabs}       
\usepackage{amsfonts}       
\usepackage{nicefrac}       
\usepackage{microtype}      
\usepackage{xcolor}         

\usepackage{threeparttable}
\usepackage{multirow}
\usepackage{graphicx}
\usepackage{newtxmath}
\usepackage{rotating}

\usepackage{wrapfig}
\usepackage{enumitem}
\usepackage{ragged2e}
\usepackage{listings}
\usepackage[misc]{ifsym}

\definecolor{codegreen}{rgb}{0,0.6,0}
\definecolor{codegray}{rgb}{0.5,0.5,0.5}
\definecolor{codepurple}{rgb}{0.58,0,0.82}
\definecolor{backcolour}{rgb}{0.9,0.9,0.9}
\lstdefinestyle{mystyle}{
    backgroundcolor=\color{backcolour},
    keywordstyle=\color{black},
    numberstyle=\tiny\color{black},
    stringstyle=\color{black},
    commentstyle=\color{black}, 
    keywordstyle=\color{black},
    stringstyle=\color{black},
    basicstyle=\ttfamily\footnotesize, 
    basicstyle=\ttfamily\footnotesize,
    breakatwhitespace=true,
    breaklines=true,
    postbreak=\mbox{\hspace{0pt}},
    captionpos=b,
    keepspaces=true,
    numbers=none,
    numbersep=5pt,
    showspaces=false,
    showstringspaces=false,
    showtabs=false,
    tabsize=2
}
\title{IA-T2I: Internet-Augmented Text-to-Image Generation}

\author{Chuanhao Li\textsuperscript{\rm 1*},
Jianwen Sun\textsuperscript{\rm 2,5*},
Yukang Feng\textsuperscript{\rm 2,5*},
Mingliang Zhai\textsuperscript{\rm 3},
\\
\textbf{Yifan Chang}\textsuperscript{\rm 4,5},
\textbf{Kaipeng Zhang}\textsuperscript{\rm 1,5}\textsuperscript{\Letter}
\\
\\
\textsuperscript{\rm 1}Shanghai AI Laboratory
\textsuperscript{\rm 2}Nankai University
\textsuperscript{\rm 3}Beijing Institute of Technology\\
\textsuperscript{\rm 4}University of Science and Technology of China
\textsuperscript{\rm 5}Shanghai Innovation Institute
}

\begin{document}

\renewcommand{\thefootnote}{\fnsymbol{footnote}}
{\let\thefootnote\relax\footnotetext{
\noindent \hspace{-5mm}
\textsuperscript{*} Equal contribution
\ \ 
\textsuperscript{\Letter} Corresponding author: \textrm{kp\_zhang@foxmail.com}
}}

\maketitle

\begin{abstract}
Current text-to-image (T2I) generation models achieve promising results, but they fail on the scenarios where the knowledge implied in the text prompt is uncertain.
For example, a T2I model released in February would struggle to generate a suitable poster for a movie premiering in April, because the character designs and styles are uncertain to the model.
To solve this problem, we propose an Internet-Augmented text-to-image generation (IA-T2I) framework to compel T2I models clear about such uncertain knowledge by providing them with reference images.
Specifically,
an active retrieval module is designed to determine whether a reference image is needed based on the given text prompt;
a hierarchical image selection module is introduced to find the most suitable image returned by an image search engine to enhance the T2I model;
a self-reflection mechanism is presented to continuously evaluate and refine the generated image to ensure faithful alignment with the text prompt.
To evaluate the proposed framework's performance, we collect a dataset named Img-Ref-T2I, where text prompts include three types of uncertain knowledge:
(1) known but rare.
(2) unknown.
(3) ambiguous.
Moreover, we carefully craft a complex prompt to guide GPT-4o in making preference evaluation, which has been shown to have an evaluation accuracy similar to that of human preference evaluation.
Experimental results demonstrate the effectiveness of our framework, outperforming GPT-4o by about 30\% in human evaluation.
\end{abstract}

\section{Introduction}

Text-to-image (T2I) generation models, such as Stable Diffusion \cite{rombach2021highresolution}, ControlNet \cite{zhang2023adding} and FLUX \cite{flux2024}, 
have attracted considerable attention for their ability to generate highly realistic images based on text prompts that often encapsulate complex and context-dependent knowledge.
However, knowledge is unevenly distributed across the world, constantly evolving, and often ambiguous.
These characteristics make it challenging for T2I models to perform reliably in scenarios where the knowledge implied in the text prompt is uncertain.
For example, a T2I model released in February would likely struggle to generate an appropriate poster for a movie premiering in April, as it lacks access to up-to-date information about the movie.
Key visual elements such as character designs, costumes, and stylistic choices may not be publicly available or finalized at the time the model was trained, which can lead to inaccurate or overly generic results.
To address this issue, we propose an Internet-Augmented Text-to-Image (IA-T2I) generation framework, which augments T2I models’ understanding of uncertain knowledge by supplying them with relevant reference images retrieved from Internet.

\begin{figure}[t]
    \vspace{-0.8cm}
    \centering
    \includegraphics[width=0.98\linewidth]{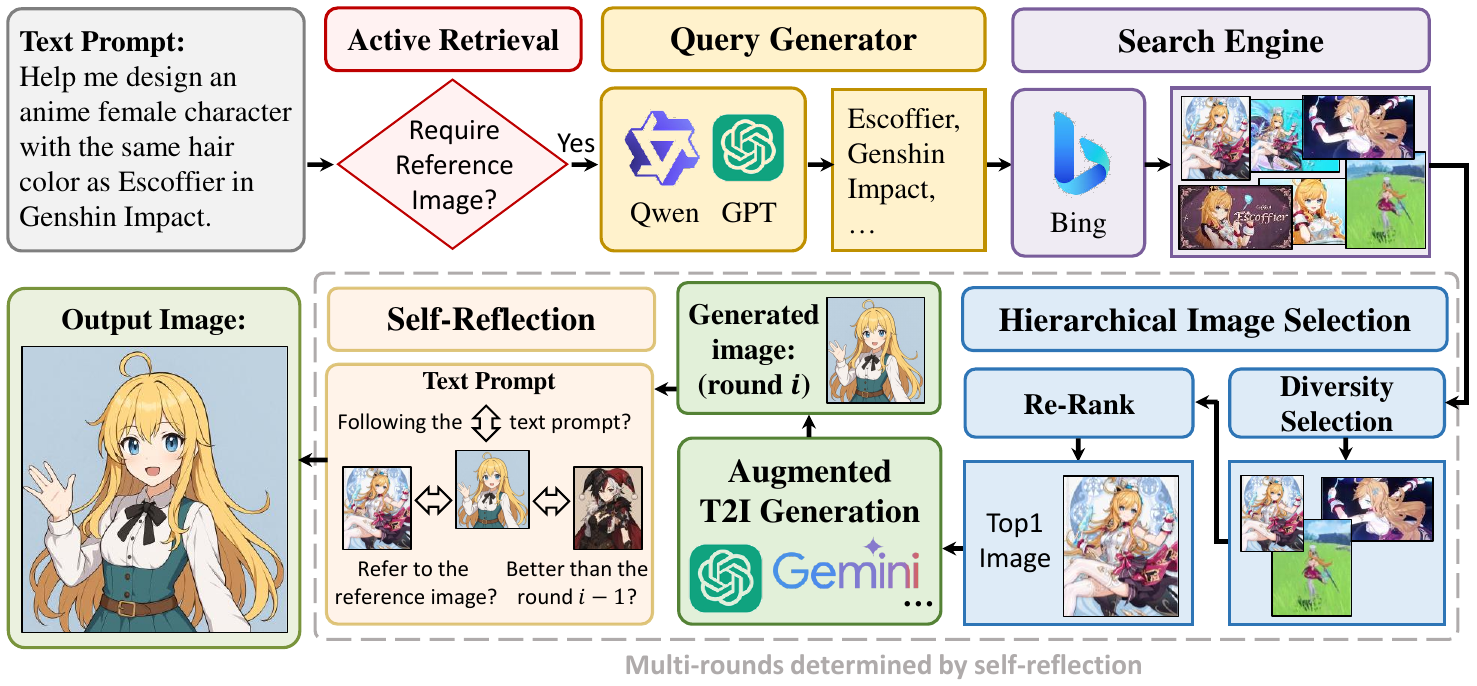}
    \caption{The proposed IA-T2I, a framework for T2I models to refer to images.}
    \label{fig:framework}
    \vspace{-0.4cm}
\end{figure}

We first present our overall framework for augmenting T2I models with reference images,
consists of six components: active retrieval module, query generator, search engine, hierarchical image selection module, augmented T2I generation, and self-reflection mechanism, as illustrated in Figure \ref{fig:framework}.
Specifically,
We begin by exploring the knowledge boundaries of a T2I model to determine whether generating an accurate image for a given text prompt requires additional reference images.
Next, we use large vision-language models (LVLMs) to extract queries from the text prompt and retrieve potentially useful reference images via search engines.
However, directly augmenting T2I models with the retrieved images is impractical due to:
(1) The number of retrieved images is typically large, making processing computationally expensive and time-consuming.
(2) Although the images are ranked by relevance, those at the top are not necessarily the most suitable as reference images.
To address this, we introduce a hierarchical image selection module to identify the most helpful reference image for guiding image generation. It first performs an initial filtering based on diversity to form a candidate set, and then re-ranks the candidates to select the most relevant and informative reference image.
Once the T2I model generates an image augmented by the selected reference image, a self-reflection mechanism is employed to evaluate the output and autonomously decide whether to reselect reference images for
another generation attempt or to accept the result as final.

We then construct a dataset dubbed Img-Ref-T2I, where text prompts include three types of uncertain knowledge:
(1) known but rare.
(2) unknown.
(3) ambiguous.
\textit{Known but rare} refers to knowledge that existed before the model's release (set as March 26, 2025, the latest update date of GPT-4o \cite{hurst2024gpt} in this paper) but is not commonly encountered.
For example, people are generally more familiar with checkers than with Polish draughts.
\textit{Unknown} refers to new knowledge that emerges after the model's release, which is inaccessible to the model.
\textit{Ambiguous} refers to a concept that has different meanings and visual representations depending on the context. For example, drawing a map of EU member states requires specifying a point in time, as the composition of the European Union has changed over the years.
Our Img-Ref-T2I dataset is designed for two tasks: general text-to-image generation and text-conditioned image editing (TI2I).
To assess the quality of generated images, we design a complex prompt to perform automatic preference evaluation pipeline based on GPT-4o, which achieves evaluation accuracy comparable to human preference evaluation.
Experimental results on the Img-Ref-T2I dataset validate the effectiveness of the proposed IA-T2I framework, outperforming GPT-4o by approximately 30\% in human evaluations.

Our contributions are summarized as follows.
(1) We propose IA-T2I, the first framework that integrates reference images from the Internet into T2I models, effectively mitigating inaccurate image generation caused by uncertain knowledge in text prompts.
(2) We collect Img-Ref-T2I, the first dataset for evaluating the performance of T2I models under three types of scenarios where the textual knowledge is uncertain.
(3) We develop a GPT-4o-based automatic preference evaluation method by prompt engineering, achieving results comparable to human preference evaluation.

\section{Related Work}

\subsection{Text-to-Image Generation}

With the continuous advancement of technologies such as generative adversarial networks \cite{goodfellow2020generative, xu2018attngan}, diffusion models \cite{song2020denoising, rombach2021highresolution, zhang2023adding}, and autoregressive models \cite{tian2024visual, sun2024autoregressive}, text-to-image generation has garnered increasing attention.
Works \cite{zhang2023adding, li2024controlnet++, liu2024lumina} emphasize the role of control signals in guiding the image generation process.
Works \cite{xie2024show, team2024chameleon, wu2024next, ge2024seed, sun2025armor} unify generation and understanding within a single model.
Works \cite{huang2024chatdit, wang2023images, wang2023seggpt, wang2023context, sun2024generative, najdenkoska2024context, huang2024context} explore the in-context learning capabilities of text-to-image models.
In contrast, we focus on retrieving helpful reference images for the image generation process to mitigate the uncertain textual knowledge, which is orthogonal to existing works.
Works \cite{blattmann2022retrieval, sheynin2022knn, chen2022re, yuan2025finerag, lyu2025realrag} explore retrieving additional reference images for text-to-image generation, where the retrieval modules require model-specific training and typically retrieve from fixed, carefully curated local databases.
In contrast, our proposed IA-T2I framework is training-free and retrieves reference images from the Internet, which is a constantly evolving and highly noisy source.

\subsection{Internet-Augmented Generation}

Recently Internet-augmented generation (IAG) attracted increasing attention of both the natural language processing (NLP) and vision-and-language (V\&L).
In NLP, 
Komeili \emph{et.al.} \cite{komeili2021internet} demonstrate that incorporating search engines into large language models (LLMs) can reduce the generation of factually incorrect content during human dialogues.
Lazaridou \emph{et.al.} \cite{lazaridou2022internet} employs few-shot prompting to allow LLMs to leverage knowledge retrieved from search engine to respond to questions involving factual and up-to-date information.
Tian \emph{et.al.} \cite{tian2023chatplug} use IAG to build open-domain generative dialogue system for digital human.
In V\&L,
Li \emph{et al.} \cite{lisearchlvlms} propose SearchLVLMs, a framework that enables existing LVLMs to access up-to-date knowledge during inference through IAG.
Jiang \emph{et al.} \cite{jiang2024mmsearch} present MMSearch to empower LVLMs for multimodal searching via IAG.
Differently, we introduce IAG into the text-to-image generation task, to mitigate inaccurate image generation caused by uncertain knowledge in text prompts.

\section{IA-T2I Framework}

In this section, we introduce IA-T2I, a framework that addresses the issue brought by uncertain knowledge by integrating reference images from the Internet into them.
The overview of IA-T2I is illustrated in Figure \ref{fig:framework}.
For a text prompt $T$ (for TI2I, the input information additionally includes an original image $I_0$), we first use an active retrieval module to determine whether a reference image is required.
Then we extract queries for $T$ and fed them into search engine to obtain potential reference images.
Next, a hierarchical image selection module is used to identify the most helpful reference image, and the T2I model is augmented by the reference image to generate an output image.
Finally, we employ a self-reflection mechanism to evaluate the output and determine whether to reselect reference images for another generation attempt or accept the result as final.

\subsection{Active Retrieval}

To determine whether a reference image is required for $T$, it is essential to explore the knowledge boundaries of a T2I model.
We attempt two different lines:
(1) Judge solely based on the input information ($T$ for T2I, $T$ and $I_0$ for TI2I).
(2) Judge based on both the input information and the image generated by the T2I model (without reference images).
For the former, we prompt the model to analyze the input information and determine whether it contains uncertain knowledge.
For the latter, we assess the instruction-following ability of the generated image to decide whether reference images are needed for regeneration.
The effectiveness of active retrieval depends on the design of the prompt.
Detailed experimental analysis and prompt choice are provided in Section~\ref{sec:active}.

\subsection{Query Generator}

We leverage existing LVLMs, such as GPT-4o \cite{hurst2024gpt} and Qwen2.5-VL \cite{bai2025qwen2}, to extract queries from $T$ to obtain queries that lead search engines to return potential helpful reference images.
Thanks to their language understanding capabilities, LVLMs can infer the grammatical role of each word in $T$, even when the knowledge implied in certain words is uncertain.
Since some knowledge is culture-specific, we translate $T$ into two additional languages including Chinese and Japanese, and extract queries accordingly to ensure a higher recall of reference images returned by search engines.
The prompt used for guiding LVLMs in generating queries
can be found in the Appendix.

\subsection{Search Engine}

The queries in three different languages are fed separately to the search engine.
By invoking the image search function, the engine directly returns multiple images.
However, directly using all the returned images to enhance the T2I model is impractical.
Firstly, many of the images are noisy and may mislead the T2I generation process.
Additionally, the sheer number of images makes it computationally expensive and time-consuming to process them all.
Although the search engine ranks the returned images based on query relevance, the top-ranked images may only match the content of the webpage they are embedded in rather than the actual visual content.
Therefore, further filtering of the returned images is necessary.

\subsection{Hierarchical Image Selection}

To filter the returned images,
we propose a hierarchical image selection module that performs two-step filtering, consisting of a diversity-based selection followed by a re-ranking process.

\textbf{Diversity Selection.}
We first perform an initial filtering of the returned images by selecting those with the greatest diversity, aiming to ensure that the candidate set includes as many useful reference images as possible.
Specifically, we extract CLIP features \cite{clip} from the returned images and apply k-means clustering \cite{jain1988algorithms} based on cosine similarity between these features to form N clusters.
The image closest to the center of each cluster is then selected as a candidate reference image.

\textbf{Re-Rank.}
The purpose of re-rank is to sort the candidate reference images and select the most helpful one.
We input multiple images into LVLMs and use the prompt (can be found in the Appendix) to guide them in ranking.
Notably, due to the self-reflection mechanism (described in Section \ref{sec:reflection}), the hierarchical image selection module may be executed in multiple rounds.
Therefore, we use $I_{i}^{ref}$ to denote the top-1 reference image selected for round $i$.

\subsection{Augmented Generation}

After obtaining the reference image $I_{i}^{ref}$,
We provide it during the image generation process of T2I models and indicate its role in the text prompt to perform augmented generation.
The output image by augmented generation of round $i$ is denoted as $I_{i}^{o}$.

\subsection{Self-Reflection}
\label{sec:reflection}

A self-reflection mechanism is employed to evaluate the accuracy and usability of the output image $I_{i}^{o}$ generated in the current round.
If the result is deemed unsatisfactory, a new round is initiated to reselect reference images and attempt generation again.
The evaluation of $I_{i}^{o}$ is based on three key criteria:
(1) Whether it faithfully follows the text prompt $T$.
(2) Whether the reference image $I_{i}^{ref}$ is helpful.
(3) Whether it effectively incorporates information from the reference image $I_{i}^{ref}$ selected in this round.
(4) For $i > 1$, whether it improves upon the output image from the previous round, \textit{i.e.}, $I_{i-1}^{o}$.
For these four criteria, we use different prompts (provided in the Appendix) to guide GPT-4o in scoring $I_{i}^{o}$.
When the total score is greater than or equal to $8$, $I_{i}^{o}$ is accepted as the final output image, denoted as $I_{out}$.

\begin{figure}[t]
    \centering
    \includegraphics[width=0.92\linewidth]{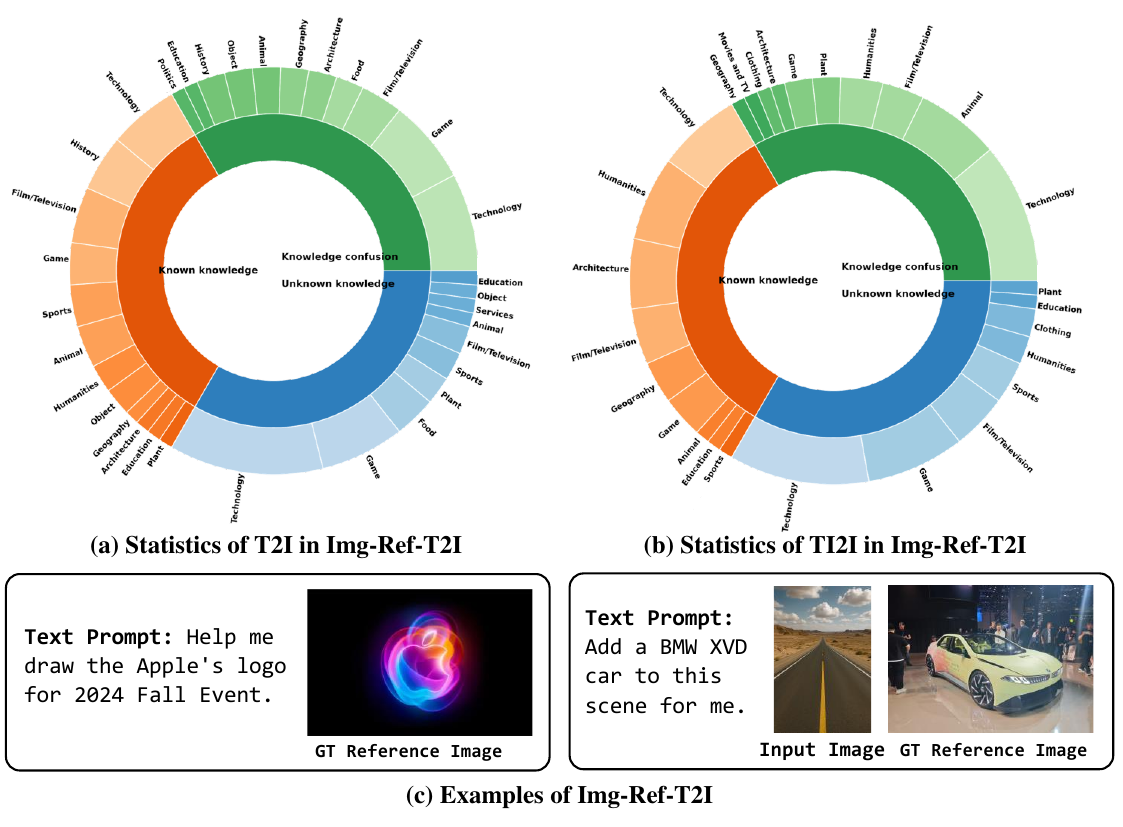}
    \caption{The distribution and examples of the proposed Img-Ref-T2I dataset.}
    \label{fig:dataset}
\end{figure}

\section{Img-Ref-T2I Dataset}

To evaluate the performance of T2I models in scenarios with uncertain textual knowledge, we construct a dataset named Img-Ref-T2I, which maintains high quality, as it is entirely curated and annotated by human experts.
The dataset consists of a total of 240 samples: 120 for T2I and another 120 for TI2I.
We categorize uncertain knowledge into three types:
(1) \textit{Known but rare},
(2) \textit{Unknown}, and
(3) \textit{Ambiguous}.
For each task’s 120 samples, 30 samples correspond to each of the three uncertainty categories.
In addition, we collect 30 samples that contain no uncertain knowledge, which are used to evaluate the accuracy of the active retrieval module.
\textit{Known but rare} refers to knowledge that existed before the release date of the model. In this paper, we define this as prior to GPT-4o’s most recent update, March 26, 2025.
\textit{Unknown} refers to new concepts or events that emerged after the model’s release.
\textit{Ambiguous} refers to concepts that exhibit different visual representations depending on the context.
For each sample containing uncertain knowledge, we manually collect a ground-truth (GT) reference image to evaluate the performance bottleneck in image generation.
The distribution and examples of the Img-Ref-T2I dataset are illustrated in Figure \ref{fig:dataset}.

\section{Experiments}

\subsection{Settings}

\textbf{Baselines.}
We select three categories of models as baselines for our experiments:
(1) T2I models, including FLUX \cite{flux2024} and DDPM \cite{huberman2024edit}.
FLUX is a commonly used T2I model, while DDPM is an inverse-based T2I model capable of injecting reference image information into the generation process.
(2) TI2I models, represented by Step1X-Edit \cite{liu2025step1x-edit}.
(3) Omnipotent (Omni.) models, including Gemini-2.0-flash \cite{gemini2} (represented by Gemini in the following text for convenience) and GPT-4o \cite{hurst2024gpt}.
These commercial models support both T2I and TI2I, and can incorporate reference images as contextual input for image generation or editing.

\textbf{Implementation Details.}
For open-source models: FLUX, DDPM, and Step1X-Edit, we re-implement them via their official code repositories.
For closed-source commercial models, GPT-4o and Gemini, we access the models via their official APIs.
We incorporate GPT-4o and Gemini into IA-T2I, as they support reference images.
We perform evaluations with a single Nvidia A100 GPU.

\subsection{Generated Image Evaluation}
\label{sec:gie}

We evaluate generated images on the proposed Img-Ref-T2I dataset via human evaluation.
For a text prompt $T$ in the T2I task, the image generated by model $X$ is denoted as $I_{out}$. Each record is represented as $[T, I_{out}^{X}]$ (for the TI2I task, the original image $I_0$ is additionally included in the record). We ask evaluators (co-authors of this paper) to score the outputs based on three aspects:
(1) Aesthetic Quality (AQ). Evaluators are asked three questions:
a. (Layout) Is the layout of $I_{out}^{X}$ harmonious? 
b. (Color) Are the colors in $I_{out}^{X}$ coordinated? 
c. (Clearness) Is $I_{out}^{X}$ visually clear? 
(2) Commonsense Consistency (CC). Evaluators are asked: ``Do the details in $I_{out}^{X}$ align with human commonsense?'' For example, an image of a cat with two tails would violate commonsense.
(3) Instruction Following (IF). Evaluators are asked: ``Does $I_{out}^{X}$ follow $T$ (or $T$ and $I_0$ in the case of TI2I)?''
All questions in the evaluation process are binary: answers must be either ``Yes'' or ``No''.
A ``Yes'' earns 1 point; ``No'' earns 0.
An image scoring 5 points is considered correct in overall.
Note that throughout the human evaluation process, model $X$ remains hidden from evaluators to ensure fairness.
Each record is evaluated by three different annotators to reduce human bias.
For each model, we normalize the scores for each evaluation criterion by dividing the total score for that criterion by the number of samples. The resulting value is used as the final score for that evaluation criterion.

Experimental results are listed in Table \ref{tab:sota}.
We can observe that:
(1) On our proposed Img-Ref-T2I dataset, existing models struggle to generate correct images, indicating that knowledge uncertainty significantly impacts the robustness of image generation.
(2) Even when using ground-truth GT reference images as context, the generation accuracy of Gemini and GPT-4o still leaves room for improvement.
(3) By leveraging the proposed IA-T2I framework, Gemini and GPT-4o achieve performance comparable to that with GT reference images, demonstrating the effectiveness of our framework in selecting appropriate reference images.

\begin{table}[tb]
    \small
    \centering
    \caption{Comparision with SOTA image generation models on Img-Ref-T2I,
    where ``Raw'' represents the model without our framework,
    ``Ours'' stands for incorporating the Raw baseline into our framework.
    ``N/A''/``GT''/``Search'' denote use ``nothing''/``ground-truth''/``our frame's'' reference images.
    The value before/after ``/" indicates the score on the T2I/TI2I task.
    }
    \setlength{\tabcolsep}{0.35mm}{
    \begin{tabular}{lclccccccccc}
        \hline\noalign{\smallskip}
         \multirow{2.5}{*}{Type} & \multirow{2.5}{*}{Model} & \multirow{2.5}{*}{Variant} & \multicolumn{3}{c}{Reference Image}  & \multicolumn{3}{c}{AQ} & \multirow{2.5}{*}{CC} & \multirow{2.5}{*}{IF} & \multirow{2.5}{*}{Overall} \\
         \cmidrule(lr){4-6}\cmidrule(lr){7-9}
         & & & N/A & GT & Search & Layout & Color & Clearness & & \\
         \noalign{\smallskip}\hline\noalign{\smallskip}
         \multirow{2}{*}{T2I}   & \multirow{1}{*}{FLUX \cite{flux2024}}     & Raw           & $\checkmark$ &      -       &      -       & 95.3/- & 93.3/- & 87.3/- & 91.3/- & 3.3/- & 3.3/- \\
                                & \multirow{1}{*}{DDPM \cite{huberman2024edit}}     & Raw           &      -       & $\checkmark$ &      -       & 77.2/- & 84.8/- & 48.7/- & 81.6/- & 72.1/- & 40.5/- \\
         \noalign{\smallskip}\hline\noalign{\smallskip}
         \multirow{1}{*}{TI2I}  & \multirow{1}{*}{Step1X-Edit \cite{liu2025step1x-edit}}   & Raw           & $\checkmark$ &      -       &      -       & -/71.9 & -/85.0 & -/48.8 & -/80.6 & -/5.0 & -/3.1 \\
         \noalign{\smallskip}\hline\noalign{\smallskip}
         \multirow{6.5}{*}{Omni.} & \multirow{3}{*}{Gemini} & Raw           & $\checkmark$ &      -       &      -       & 96.3/91.2 & 97.5/96.0 & 88.2/89.3 & 91.9/91.9 & 12.9/9.4 & 11.1/6.7 \\
                                &                                   & Raw           &      -       & $\checkmark$ &      -       & 93.7/93.0 & 96.2/95.5 & 81.8/87.3 & 92.5/95.0 & 67.3/31.2 & 56.6/26.1 \\
                                &                                   & \textbf{Ours} &      -       &      -       & $\checkmark$ & 95.8/87.7 & 97.0/96.7 & 90.4/83.2 & 94.6/89.7 & 52.1/13.5 & 45.5/10.3 \\
         \cmidrule(l){2-12}
                                & \multirow{3}{*}{GPT-4o}   & Raw           & $\checkmark$ &      -       &      -       & 98.0/96.3 & 99.3/100 & 98.0/98.1 & 97.4/94.4 & 20.4/33.3 & 19.1/32.7 \\
                                &                                   & Raw           &      -       & $\checkmark$ &      -       & 99.3/97.4 & 100/100 & 98.5/96.8 & 96.3/96.8 & 69.9/76.0 & 65.4/72.7 \\
                                &                                   & \textbf{Ours} &      -       &      -       & $\checkmark$ & 98.6/98.6 & 99.3/100 & 93.7/95.7 & 95.8/96.4 & 55.2/61.2 & 48.3/57.6 \\
         \hline\noalign{\smallskip}
    \end{tabular}}
    \label{tab:sota}
\end{table}

\subsection{Preference Evaluation}

\begin{figure}[t]
    \centering
    \includegraphics[width=1\linewidth]{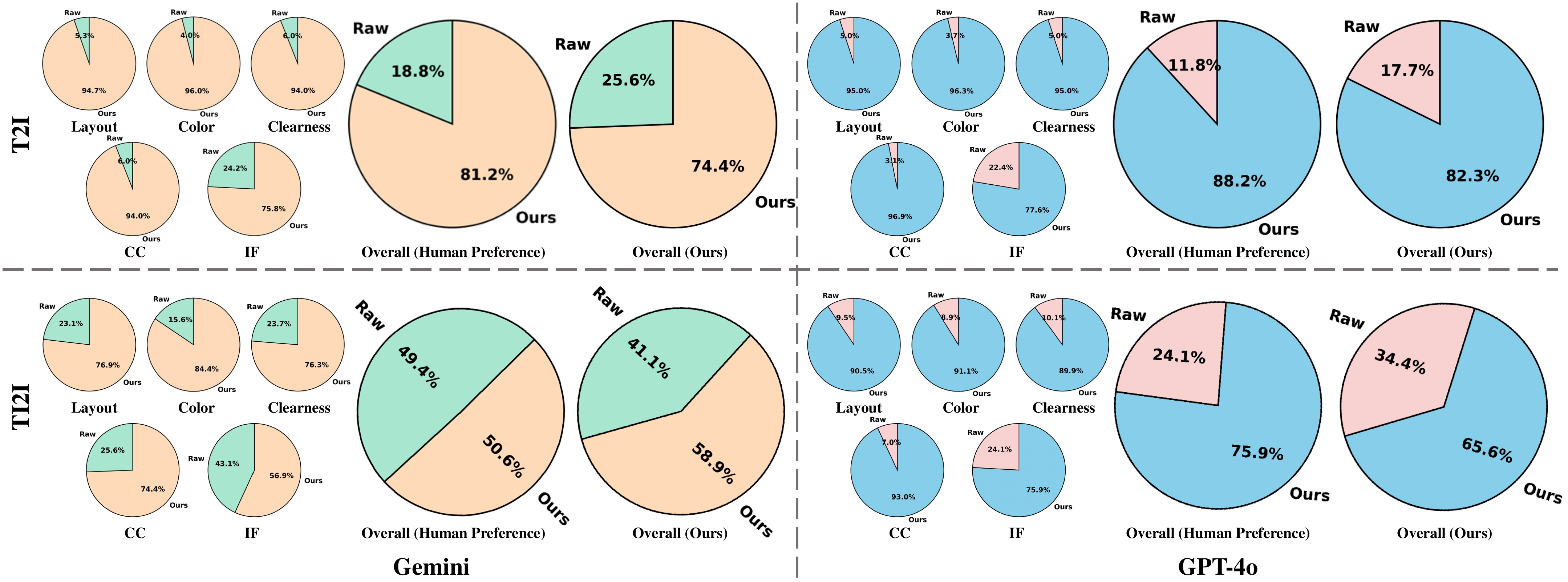}
    \caption{Experimental results of preference evaluation.}
    \label{fig:preference}
\end{figure}

Preference evaluation refers to the process of comparing two images, $I_{out}^{X_1}$ and $I_{out}^{X_2}$, generated by models $X_1$ and $X_2$ respectively for a given text prompt $T$. By assessing $I_{out}^{X_1}$ and $I_{out}^{X_2}$ from multiple dimensions, we determine which model performs better.

\textbf{Human Preference Evaluation.}
Similar to Section \ref{sec:gie}, we begin the human evaluation by comparing $I_{out}^{X_1}$ and $I_{out}^{X_2}$ from three aspects: aesthetic quality, commonsense consistency, and instruction alignment. However, unlike Section \ref{sec:gie}, the evaluation questions shift from ``Is it …?'' to ``Which model, $X_1$ or $X_2$, performs better?'' In addition, we introduce an extra question: ``Considering all the above aspects, the image generated by which model is more suitable as the final generated result?'' to evaluate the overall preference. All questions are designed to have binary answers, either ``$X_1$'' or ``$X_2$'', to prevent ambiguous or indecisive responses from evaluators.

We conduct the human preference evaluation by comparing the baseline models with their counterparts integrated into our proposed framework, \textit{i.e.}, $X_1$ and $X_2$, respectively.
Two sets of evaluations were performed, with the baseline models being Gemini and GPT-4o. The experimental results are presented in Figure~\ref{fig:preference}.
It can be observed that, for both the T2I and TI2I tasks, human evaluators consistently prefer the images generated using our framework.
This indicates that the proposed framework significantly improves the performance of the baseline models.

\textbf{GPT-4o Preference Evaluation.}
Conducting large-scale human preference evaluations is extremely costly.
To address this, we design a complex prompt to guide GPT-4o in making preference evaluation based on GPT-4o, which enables automatic assessment and significantly reduces human labor costs.
For the T2I task, GPT-4o preference evaluation relies on the input set [$T$, $I_1$, $I_2$, $I_\text{ref}$].
For the TI2I task, it additionally considers the initial image $I_0$.
The prompt is provided in the Appedix.
Experimental results of this automated preference evaluation are denoted as Ours and shown in Figure~\ref{fig:preference}, demonstrating that the scores produced by our pipeline are comparable to those from the human preference evaluation, highlighting the potential of automatic preference evaluation.

\begin{table}[ht]
        \small
	\centering
  	\caption{Ablation studies of our framework on Img-Ref-T2I.}
	\setlength{\tabcolsep}{0.8mm}{
	\begin{tabular}{clcccccccccc}
	\hline\noalign{\smallskip}
        \multirow{2.5}{*}{Model} & \multirow{2.5}{*}{Variant} & \multicolumn{3}{c}{Query Generator} & \multirow{1.5}{*}{Diversity} & \multicolumn{3}{c}{Re-Rank} & \multirow{1.5}{*}{Self} & \multicolumn{2}{c}{Acc.} \\
	\cmidrule(lr){3-5}\cmidrule(lr){7-9}\cmidrule(lr){11-12}
        &  & \textit{Ori.}  & \textit{Qwen2.5-VL} & \textit{GPT-4o}  &  \multirow{0.5}{*}{Selection} & \textit{Qwen2.5-VL} & \textit{GPT-4o} & \textit{Human}  &  \multirow{0.5}{*}{Reflection} & T2I & TI2I\\
		\noalign{\smallskip}\hline\noalign{\smallskip}
        \multirow{8.5}{*}{\rotatebox[origin=c]{90}{GPT-4o}} & Raw    &       -      &      -       &      -       &      -       &      -       &      -        &      -       & - & 19.1 & 32.7 \\
        \cmidrule(l){2-12}
        & \multirow{7}{*}{\textbf{Ours}}      & $\checkmark$ &      -       &      -       &      -       &      -       &      -       & $\checkmark$  &      -       & 51.4 & 61.2 \\
                                        &     &      -       & $\checkmark$ &      -       &      -       &      -       &      -       & $\checkmark$  &      -       & 56.1 & 67.6 \\
                                        &     &      -       &      -       & $\checkmark$ &      -       &      -       &      -       & $\checkmark$  &      -       & 58.0 & 69.7 \\
                                        &     &      -       &      -       & $\checkmark$ &      -       & $\checkmark$ &      -       &      -        &      -       & 25.8 & 39.8 \\
                                        &     &      -       &      -       & $\checkmark$ &      -       &      -       & $\checkmark$ &      -        &      -       & 26.9 & 40.2 \\
                                        &     &      -       &      -       & $\checkmark$ & $\checkmark$ &      -       & $\checkmark$ &      -        & - & 40.8 & 49.4 \\
                                        &     &      -       &      -       & $\checkmark$ & $\checkmark$ &      -       & $\checkmark$ &      -        & $\checkmark$ &  \textbf{48.3} & \textbf{57.6} \\
		\noalign{\smallskip}\hline
	\end{tabular}}
	\label{tab:ablation}
\end{table}

\vspace{-0.2cm}
\subsection{Ablation Studies}

The results of the ablation study are shown in Table~\ref{tab:ablation}, where ``Acc.'' denotes the overall score of the generated images as assessed by human evaluation.
We incrementally integrate different components of our proposed framework on top of GPT-4o, and explored several implementation variants for specific modules.
The following observations can be made:
(1) The framework is not highly sensitive to the choice of query generator. Using the original text prompt as the query (Ori.) or extracting queries via Qwen2.5-VL \cite{bai2025qwen2} yields competitive performance.
(2) When reference images are manually selected by humans (Human), the generated images are still not entirely accurate, indicating that T2I and TI2I models require further improvements in utilizing reference images effectively.
(3) Performing an initial filtering of search engine results using image clustering significantly enhances generation accuracy by reducing the difficulty of the re-ranking process.
(4) Incorporating a self-reflection mechanism further improves generation accuracy.
These findings demonstrate that the components of our framework are both effective and complementary.

\begin{wraptable}{r}{7cm}
\vspace{-0.5cm}
        \small
	\centering
  	\caption{Ablation studies of self-reflection.}
	\setlength{\tabcolsep}{0.5mm}{
	\begin{tabular}{clccccc}
	\hline\noalign{\smallskip}
        \multirow{1}{*}{Model} & \multirow{1}{*}{Variant} & \textit{$(I_i^o, T)$}  & \textit{$(I_i^o, I_i^{ref})$} & \textit{$(I_i^o, I_{i-1}^o)$} & T2I & TI2I \\
		\noalign{\smallskip}\hline\noalign{\smallskip}
        \multirow{4.5}{*}{\rotatebox[origin=c]{90}{GPT-4o}} & Raw    &       -      &      -       &      -       & 19.1 & 32.7 \\
        \cmidrule(l){2-7}
        & \multirow{3}{*}{\textbf{Ours}}      & $\checkmark$ &      -       &      -       & 41.7 & 46.6 \\
                                        &     & $\checkmark$ & $\checkmark$ &      -       & 43.6 & 51.2 \\
                                        &     & $\checkmark$ & $\checkmark$ & $\checkmark$ & \textbf{48.3} & \textbf{57.6} \\
		\noalign{\smallskip}\hline
	\end{tabular}}
	\label{tab:ablation_sr}
\end{wraptable}

We perform an ablation study on the criteria used in the self-reflection mechanism, with the results shown in Table~\ref{tab:ablation_sr}.
\textit{$(I_i^o, T)$}, \textit{$(I_i^o, I_i^{ref})$}, and \textit{$(I_i^o, I_{i-1}^o)$} correspond to criteria 1, 2, and 3 in Section \ref{sec:reflection}, respectively.
The results indicate that employing all three criteria together leads to higher image generation accuracy, demonstrating not only the complementarity of the criteria but also the rationality and effectiveness of the self-reflection mechanism.

\subsection{Analysis of Active Retrieval}
\label{sec:active}

\begin{wraptable}{r}{5cm}
        \vspace{-0.6cm}
        \small
	\centering
  	\caption{Analysis of active retrieval.}
	\setlength{\tabcolsep}{1.5mm}{
	\begin{tabular}{cccccc}
	\hline\noalign{\smallskip}
        \multirow{2.5}{*}{Task} & \multirow{2.5}{*}{Prompt} & \multicolumn{3}{c}{Dependency} & \multirow{2.5}{*}{Acc.} \\
	\cmidrule(lr){3-5}
        &  & $T$ & $I^o$ & $I_0$ & \\
		\noalign{\smallskip}\hline\noalign{\smallskip}
        \multirow{3}{*}{T2I} & prompt1 & $\checkmark$ & - & - & 90.0 \\
                             & prompt2 & $\checkmark$ & - & - & 80.8 \\
                             & prompt3 & $\checkmark$ & $\checkmark$ & - & \textbf{95.8} \\
        \cmidrule(l){1-6}
        \multirow{3}{*}{TI2I} & prompt6 & $\checkmark$ & -  & - & \textbf{91.7} \\
                              & prompt4 & $\checkmark$ & $\checkmark$ & - & 74.2 \\
                              & prompt5 & $\checkmark$ & $\checkmark$ & $\checkmark$ & 80.8 \\
		\noalign{\smallskip}\hline
	\end{tabular}}
	\label{tab:active}
\end{wraptable}

\begin{figure}[tb]
    \centering
    \includegraphics[width=0.98\linewidth]{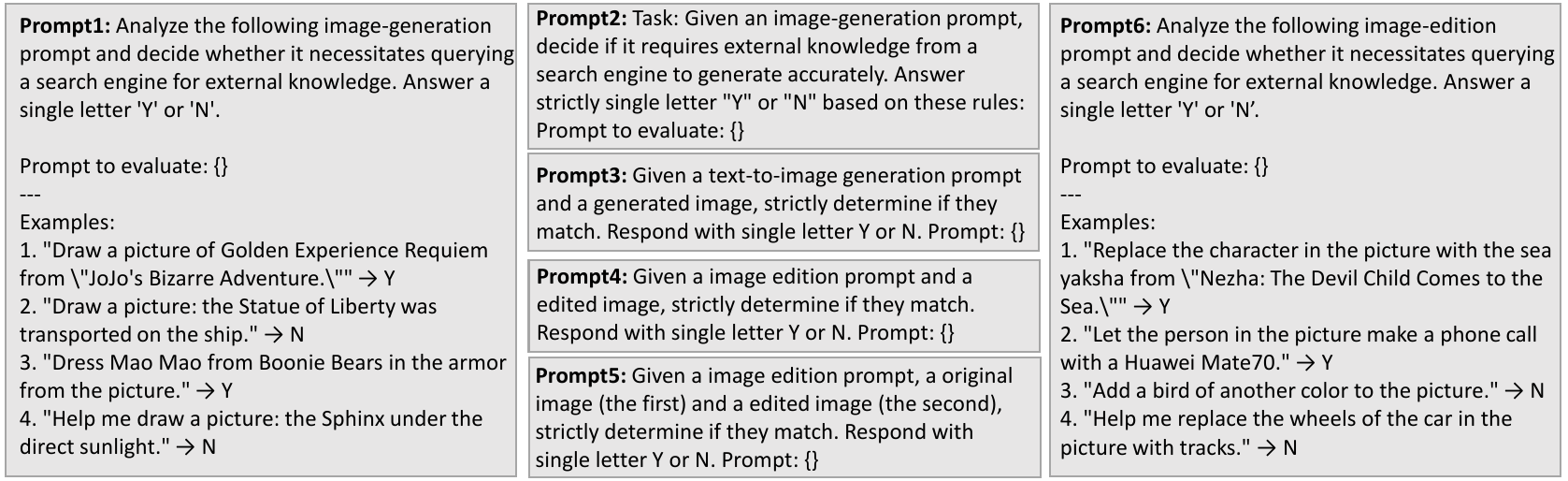}
    \caption{Prompt candidates for active retrieval.}
    \label{fig:prompt}
\end{figure}

For the two distinct lines of the active retrieval module mentioned in Section \ref{sec:active}, we conduct experiments using different prompts, and the results are shown in Table \ref{tab:active}. In this table, ``Dependency'' indicates which inputs a given prompt relies on to determine whether to trigger active retrieval. For the specific information of each prompt, please refer to Figure \ref{fig:prompt}. Specifically, $T$, $I^o$, and $I_0$ represent the text prompt, the image output by the model without reference images, and the original image (only applicable to the TI2I task), respectively. ``Acc.'' refers to the accuracy of the active retrieval module, defined as the percentage of correct decisions: for samples containing uncertain knowledge, the module should output ``Y'', and for those without, ``N''.
We observe the following:
(1) For T2I, using more dependencies and providing more detailed prompts can lead to higher accuracy.
(2) For TI2I, due to the increased complexity, relying solely on the textual input $T$ for judging active retrieval actually yields better results.
This implies that dependencies and prompts should be adjusted according to the specific task.

\begin{wrapfigure}{r}{0.4\textwidth}
    \vspace{-0.6cm}
    \centering
    \includegraphics[width=1\linewidth]{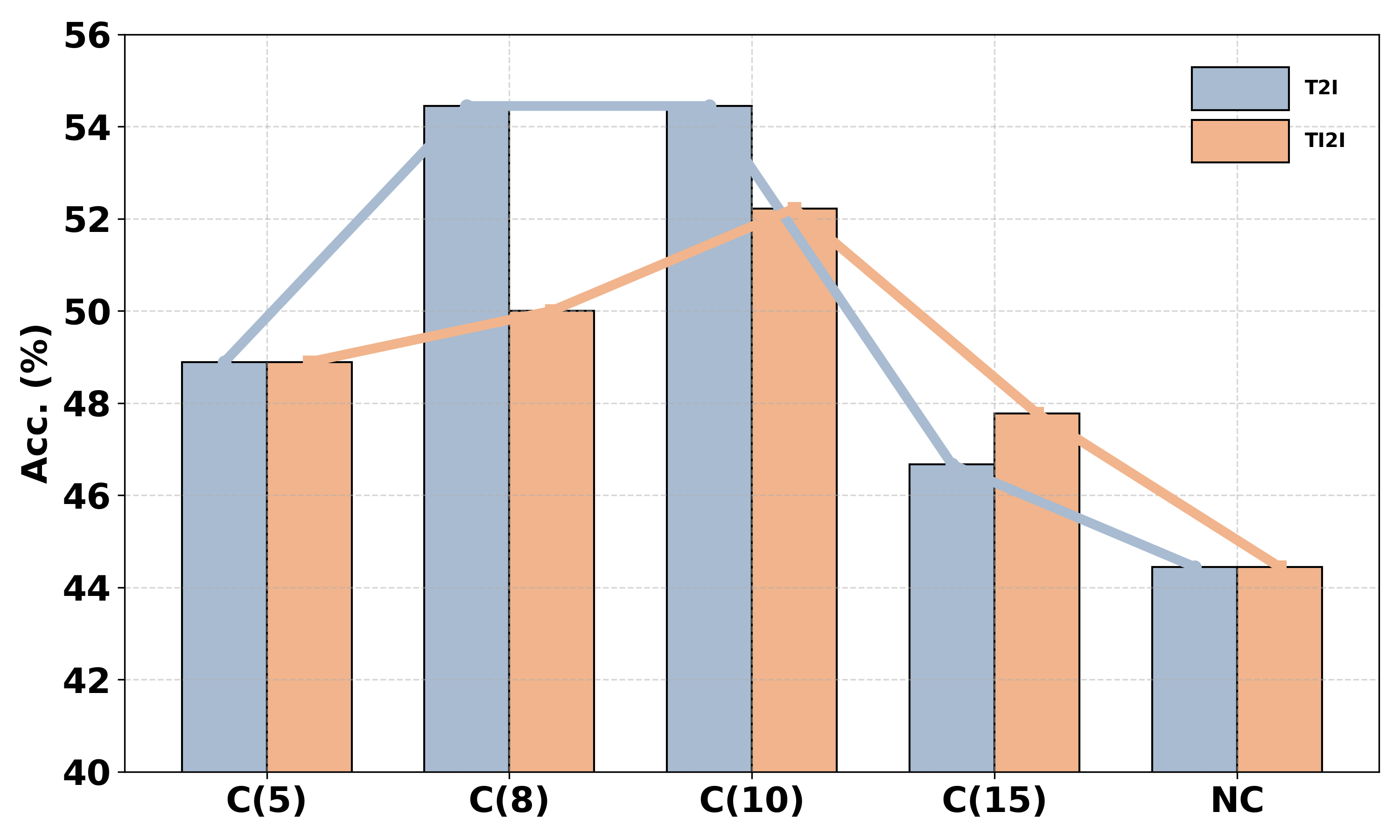}
    \caption{Analysis of diversity selection.}
    \label{fig:cluster}
    \vspace{-0.4cm}
\end{wrapfigure}

\subsection{Analysis of Diversity Selection}

This section investigates the necessity of diversity selection. We compared the performance of using different numbers of clusters for diversity selection with a baseline where all images retrieved are directly used for Re-Rank. The experimental results are shown in Figure \ref{fig:cluster}. ``Acc.'' represents the ratio of cases where the Top-1 image obtained through Re-Rank is a suitable reference image, as judged by humans.
``C($x$)'' indicates the use of diversity selection with $x$ clusters, while ``NC'' means no diversity selection is applied.
From the figure, the following observations can be made:
(1) When the number of clusters is less than 10, as the number of clusters increases, the ratio of correctly ranked Top-1 images also increases.
(2) When the number of clusters exceeds 10, the correct Re-Rank rate begins to decrease as the number of clusters increases.
(3) When no diversity selection is used, the accuracy of Re-Rank is the lowest, because an excessive number of images introduces noise into the Re-Rank process.
These observations demonstrate that diversity selection is both necessary and effective.

\vspace{-0.05cm}
\subsection{Qualitative Analysis}

\begin{figure}[tb]
    \centering
    \includegraphics[width=1\linewidth]{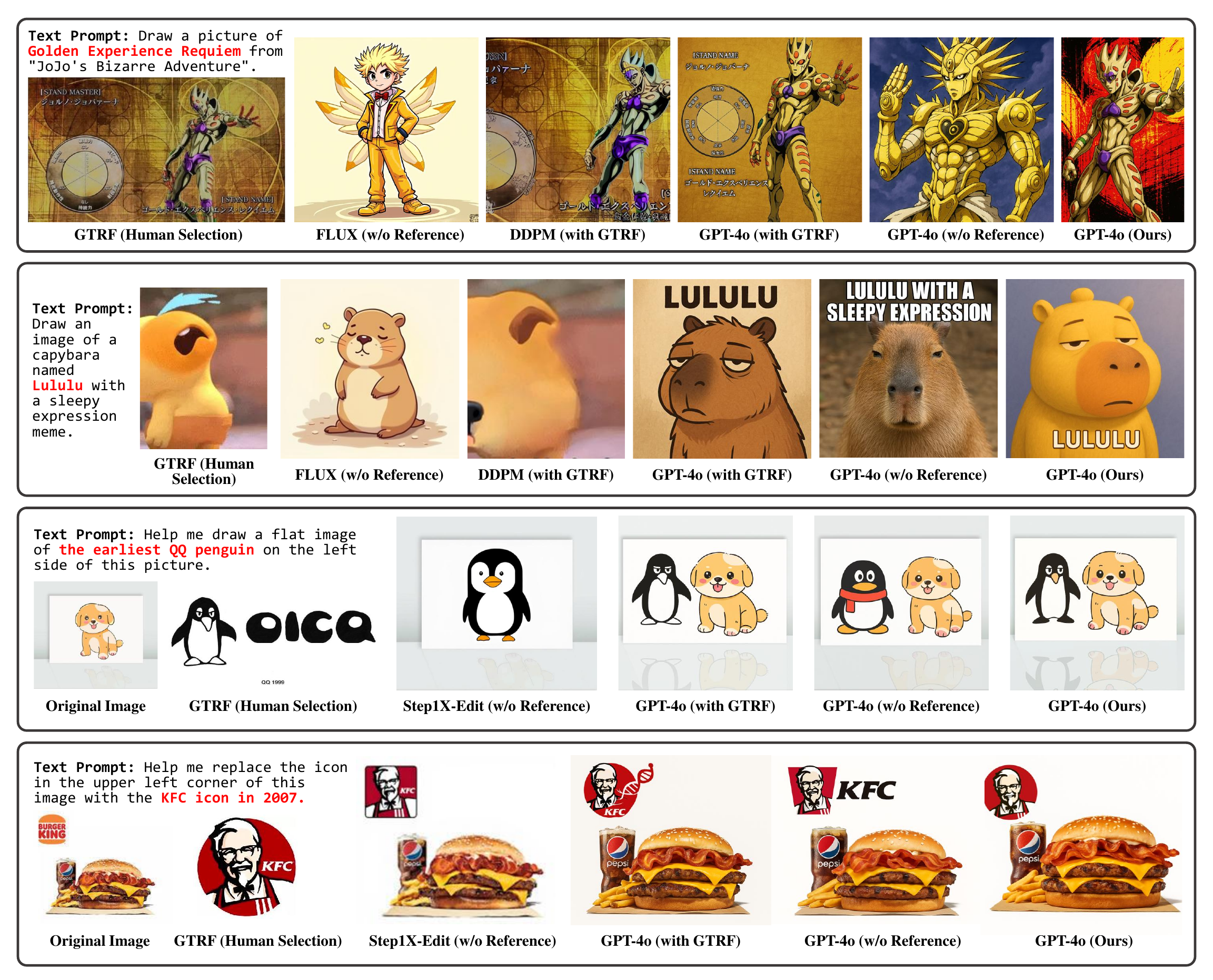}
    \caption{Qualitative comparisons on the proposed Img-Ref-T2I dataset, where GTRF denotes the ground-truth reference image provided in Img-Ref-T2I by human selection.}
    \label{fig:quality}
\end{figure}

Figure \ref{fig:quality} depicts several qualitative examples from the Img-Ref-T2I dataset of different models.
The words in bold red in the text prompt denotes a short description of the ground-truth reference image (GTRF).
For both T2I and TI2I, we provide two qualitative examples respectively.
We can observe the following:
(1) GPT-4o (Ours) performs the best, consistently generating reasonable images across all four qualitative examples.
(2) GPT-4o (with GTRF) performs slightly worse than GPT-4o (Ours).
Due to the availability of only one GTRF and the lack of a self-reflection process, the generated images often contain minor issues. For example, in the fourth example, the icon generated by GPT-4o (with GTRF) has noticeable flaws.
(3) DDPM (with GTRF) uses the GTRF but merely imitates it without deeper understanding.
(4) FLUX (w/o Reference) and Step1X-Edit (w/o Reference) cannot utilize any reference images and perform poorly on the T2I and TI2I tasks, respectively.
These observations demonstrate that the proposed IA-T2I framework is effective and, in some scenarios, can even produce results that are better than the GTRF.

\section{Limitation}

The Img-Ref-T2I dataset constructed in this paper relies on manual collection and annotation, making large-scale expansion extremely costly.
In the future, we plan to explore automatic collection and annotation of T2I and TI2I samples with uncertain knowledge.

\section{Conclusion}

In this work, we have presented IA-T2I, a framework that integrates reference images from the Internet into T2I/TI2I models, which can effectively mitigate inaccurate image generation caused by uncertain knowledge in text prompt.
An Img-Ref-T2I dataset that includes three types of scenarios involving uncertain knowledge is curated by human experts.
The dataset enables evaluate the ability of T2I/TI2I models to generate images when the textual knowledge is uncertain.
Experimental results on Img-Ref-T2I demonstrate that our framework can significantly enhances the image generation performance of T2I/TI2I models in these challenging scenarios.

\newpage
\small
\bibliographystyle{unsrt}
\bibliography{ref}

\begin{thebibliography}{10}

\bibitem{rombach2021highresolution}
Robin Rombach, Andreas Blattmann, Dominik Lorenz, Patrick Esser, and Björn Ommer.
\newblock High-resolution image synthesis with latent diffusion models, 2021.

\bibitem{zhang2023adding}
Lvmin Zhang, Anyi Rao, and Maneesh Agrawala.
\newblock Adding conditional control to text-to-image diffusion models.
\newblock In {\em Proceedings of the IEEE/CVF international conference on computer vision}, pages 3836--3847, 2023.

\bibitem{flux2024}
Black~Forest Labs.
\newblock Flux.
\newblock \url{https://github.com/black-forest-labs/flux}, 2024.

\bibitem{hurst2024gpt}
Aaron Hurst, Adam Lerer, Adam~P Goucher, Adam Perelman, Aditya Ramesh, Aidan Clark, AJ~Ostrow, Akila Welihinda, Alan Hayes, Alec Radford, et~al.
\newblock Gpt-4o system card.
\newblock {\em arXiv preprint arXiv:2410.21276}, 2024.

\bibitem{goodfellow2020generative}
Ian Goodfellow, Jean Pouget-Abadie, Mehdi Mirza, Bing Xu, David Warde-Farley, Sherjil Ozair, Aaron Courville, and Yoshua Bengio.
\newblock Generative adversarial networks.
\newblock {\em Communications of the ACM}, 63(11):139--144, 2020.

\bibitem{xu2018attngan}
Tao Xu, Pengchuan Zhang, Qiuyuan Huang, Han Zhang, Zhe Gan, Xiaolei Huang, and Xiaodong He.
\newblock Attngan: Fine-grained text to image generation with attentional generative adversarial networks.
\newblock In {\em Proceedings of the IEEE conference on computer vision and pattern recognition}, pages 1316--1324, 2018.

\bibitem{song2020denoising}
Jiaming Song, Chenlin Meng, and Stefano Ermon.
\newblock Denoising diffusion implicit models.
\newblock {\em arXiv preprint arXiv:2010.02502}, 2020.

\bibitem{tian2024visual}
Keyu Tian, Yi~Jiang, Zehuan Yuan, Bingyue Peng, and Liwei Wang.
\newblock Visual autoregressive modeling: Scalable image generation via next-scale prediction.
\newblock {\em Advances in neural information processing systems}, 37:84839--84865, 2024.

\bibitem{sun2024autoregressive}
Peize Sun, Yi~Jiang, Shoufa Chen, Shilong Zhang, Bingyue Peng, Ping Luo, and Zehuan Yuan.
\newblock Autoregressive model beats diffusion: Llama for scalable image generation.
\newblock {\em arXiv preprint arXiv:2406.06525}, 2024.

\bibitem{li2024controlnet++}
Ming Li, Taojiannan Yang, Huafeng Kuang, Jie Wu, Zhaoning Wang, Xuefeng Xiao, and Chen Chen.
\newblock Controlnet++: Improving conditional controls with efficient consistency feedback: Project page: liming-ai. github. io/controlnet\_plus\_plus.
\newblock In {\em European Conference on Computer Vision}, pages 129--147. Springer, 2024.

\bibitem{liu2024lumina}
Dongyang Liu, Shitian Zhao, Le~Zhuo, Weifeng Lin, Yu~Qiao, Hongsheng Li, and Peng Gao.
\newblock Lumina-mgpt: Illuminate flexible photorealistic text-to-image generation with multimodal generative pretraining.
\newblock {\em arXiv preprint arXiv:2408.02657}, 2024.

\bibitem{xie2024show}
Jinheng Xie, Weijia Mao, Zechen Bai, David~Junhao Zhang, Weihao Wang, Kevin~Qinghong Lin, Yuchao Gu, Zhijie Chen, Zhenheng Yang, and Mike~Zheng Shou.
\newblock Show-o: One single transformer to unify multimodal understanding and generation.
\newblock {\em arXiv preprint arXiv:2408.12528}, 2024.

\bibitem{team2024chameleon}
Chameleon Team.
\newblock Chameleon: Mixed-modal early-fusion foundation models.
\newblock {\em arXiv preprint arXiv:2405.09818}, 2024.

\bibitem{wu2024next}
Shengqiong Wu, Hao Fei, Leigang Qu, Wei Ji, and Tat-Seng Chua.
\newblock Next-gpt: Any-to-any multimodal llm.
\newblock In {\em Forty-first International Conference on Machine Learning}, 2024.

\bibitem{ge2024seed}
Yuying Ge, Sijie Zhao, Jinguo Zhu, Yixiao Ge, Kun Yi, Lin Song, Chen Li, Xiaohan Ding, and Ying Shan.
\newblock Seed-x: Multimodal models with unified multi-granularity comprehension and generation.
\newblock {\em arXiv preprint arXiv:2404.14396}, 2024.

\bibitem{sun2025armor}
Jianwen Sun, Yukang Feng, Chuanhao Li, Fanrui Zhang, Zizhen Li, Jiaxin Ai, Sizhuo Zhou, Yu~Dai, Shenglin Zhang, and Kaipeng Zhang.
\newblock Armor v0. 1: Empowering autoregressive multimodal understanding model with interleaved multimodal generation via asymmetric synergy.
\newblock {\em arXiv preprint arXiv:2503.06542}, 2025.

\bibitem{huang2024chatdit}
Lianghua Huang, Wei Wang, Zhi-Fan Wu, Yupeng Shi, Chen Liang, Tong Shen, Han Zhang, Huanzhang Dou, Yu~Liu, and Jingren Zhou.
\newblock Chatdit: A training-free baseline for task-agnostic free-form chatting with diffusion transformers.
\newblock {\em arXiv preprint arXiv:2412.12571}, 2024.

\bibitem{wang2023images}
Xinlong Wang, Wen Wang, Yue Cao, Chunhua Shen, and Tiejun Huang.
\newblock Images speak in images: A generalist painter for in-context visual learning.
\newblock In {\em Proceedings of the IEEE/CVF Conference on Computer Vision and Pattern Recognition}, pages 6830--6839, 2023.

\bibitem{wang2023seggpt}
Xinlong Wang, Xiaosong Zhang, Yue Cao, Wen Wang, Chunhua Shen, and Tiejun Huang.
\newblock Seggpt: Towards segmenting everything in context.
\newblock In {\em Proceedings of the IEEE/CVF International Conference on Computer Vision}, pages 1130--1140, 2023.

\bibitem{wang2023context}
Zhendong Wang, Yifan Jiang, Yadong Lu, Pengcheng He, Weizhu Chen, Zhangyang Wang, Mingyuan Zhou, et~al.
\newblock In-context learning unlocked for diffusion models.
\newblock {\em Advances in Neural Information Processing Systems}, 36:8542--8562, 2023.

\bibitem{sun2024generative}
Quan Sun, Yufeng Cui, Xiaosong Zhang, Fan Zhang, Qiying Yu, Yueze Wang, Yongming Rao, Jingjing Liu, Tiejun Huang, and Xinlong Wang.
\newblock Generative multimodal models are in-context learners.
\newblock In {\em Proceedings of the IEEE/CVF Conference on Computer Vision and Pattern Recognition}, pages 14398--14409, 2024.

\bibitem{najdenkoska2024context}
Ivona Najdenkoska, Animesh Sinha, Abhimanyu Dubey, Dhruv Mahajan, Vignesh Ramanathan, and Filip Radenovic.
\newblock Context diffusion: In-context aware image generation.
\newblock In {\em European Conference on Computer Vision}, pages 375--391. Springer, 2024.

\bibitem{huang2024context}
Lianghua Huang, Wei Wang, Zhi-Fan Wu, Yupeng Shi, Huanzhang Dou, Chen Liang, Yutong Feng, Yu~Liu, and Jingren Zhou.
\newblock In-context lora for diffusion transformers.
\newblock {\em arXiv preprint arXiv:2410.23775}, 2024.

\bibitem{blattmann2022retrieval}
Andreas Blattmann, Robin Rombach, Kaan Oktay, Jonas M{\"u}ller, and Bj{\"o}rn Ommer.
\newblock Retrieval-augmented diffusion models.
\newblock {\em Advances in Neural Information Processing Systems}, 35:15309--15324, 2022.

\bibitem{sheynin2022knn}
Shelly Sheynin, Oron Ashual, Adam Polyak, Uriel Singer, Oran Gafni, Eliya Nachmani, and Yaniv Taigman.
\newblock Knn-diffusion: Image generation via large-scale retrieval.
\newblock {\em arXiv preprint arXiv:2204.02849}, 2022.

\bibitem{chen2022re}
Wenhu Chen, Hexiang Hu, Chitwan Saharia, and William~W Cohen.
\newblock Re-imagen: Retrieval-augmented text-to-image generator.
\newblock {\em arXiv preprint arXiv:2209.14491}, 2022.

\bibitem{yuan2025finerag}
Huaying Yuan, Ziliang Zhao, Shuting Wang, Shitao Xiao, Minheng Ni, Zheng Liu, and Zhicheng Dou.
\newblock Finerag: Fine-grained retrieval-augmented text-to-image generation.
\newblock In {\em Proceedings of the 31st International Conference on Computational Linguistics}, pages 11196--11205, 2025.

\bibitem{lyu2025realrag}
Yuanhuiyi Lyu, Xu~Zheng, Lutao Jiang, Yibo Yan, Xin Zou, Huiyu Zhou, Linfeng Zhang, and Xuming Hu.
\newblock Realrag: Retrieval-augmented realistic image generation via self-reflective contrastive learning.
\newblock {\em arXiv preprint arXiv:2502.00848}, 2025.

\bibitem{komeili2021internet}
Mojtaba Komeili, Kurt Shuster, and Jason Weston.
\newblock Internet-augmented dialogue generation.
\newblock {\em arXiv preprint arXiv:2107.07566}, 2021.

\bibitem{lazaridou2022internet}
Angeliki Lazaridou, Elena Gribovskaya, Wojciech Stokowiec, and Nikolai Grigorev.
\newblock Internet-augmented language models through few-shot prompting for open-domain question answering.
\newblock {\em arXiv preprint arXiv:2203.05115}, 2022.

\bibitem{tian2023chatplug}
Junfeng Tian, Hehong Chen, Guohai Xu, Ming Yan, Xing Gao, Jianhai Zhang, Chenliang Li, Jiayi Liu, Wenshen Xu, Haiyang Xu, et~al.
\newblock Chatplug: Open-domain generative dialogue system with internet-augmented instruction tuning for digital human.
\newblock {\em arXiv preprint arXiv:2304.07849}, 2023.

\bibitem{lisearchlvlms}
Chuanhao Li, Zhen Li, Chenchen Jing, Shuo Liu, Wenqi Shao, Yuwei Wu, Ping Luo, Yu~Qiao, and Kaipeng Zhang.
\newblock Searchlvlms: A plug-and-play framework for augmenting large vision-language models by searching up-to-date internet knowledge.
\newblock In {\em The Thirty-eighth Annual Conference on Neural Information Processing Systems}, 2024.

\bibitem{jiang2024mmsearch}
Dongzhi Jiang, Renrui Zhang, Ziyu Guo, Yanmin Wu, Jiayi Lei, Pengshuo Qiu, Pan Lu, Zehui Chen, Chaoyou Fu, Guanglu Song, et~al.
\newblock Mmsearch: Benchmarking the potential of large models as multi-modal search engines.
\newblock {\em arXiv preprint arXiv:2409.12959}, 2024.

\bibitem{bai2025qwen2}
Shuai Bai, Keqin Chen, Xuejing Liu, Jialin Wang, Wenbin Ge, Sibo Song, Kai Dang, Peng Wang, Shijie Wang, Jun Tang, et~al.
\newblock Qwen2. 5-vl technical report.
\newblock {\em arXiv preprint arXiv:2502.13923}, 2025.

\bibitem{clip}
Alec Radford, Jong~Wook Kim, Chris Hallacy, Aditya Ramesh, Gabriel Goh, Sandhini Agarwal, Girish Sastry, Amanda Askell, Pamela Mishkin, Jack Clark, et~al.
\newblock Learning transferable visual models from natural language supervision.
\newblock In {\em Proceedings of the International Conference on Machine Learning (ICML)}, pages 8748--8763, 2021.

\bibitem{jain1988algorithms}
Anil~K Jain and Richard~C Dubes.
\newblock {\em Algorithms for clustering data}.
\newblock Prentice-Hall, Inc., 1988.

\bibitem{huberman2024edit}
Inbar Huberman-Spiegelglas, Vladimir Kulikov, and Tomer Michaeli.
\newblock An edit friendly ddpm noise space: Inversion and manipulations.
\newblock In {\em Proceedings of the IEEE/CVF Conference on Computer Vision and Pattern Recognition}, pages 12469--12478, 2024.

\bibitem{liu2025step1x-edit}
Shiyu Liu, Yucheng Han, Peng Xing, Fukun Yin, Rui Wang, Wei Cheng, Jiaqi Liao, Yingming Wang, Honghao Fu, Chunrui Han, Guopeng Li, Yuang Peng, Quan Sun, Jingwei Wu, Yan Cai, Zheng Ge, Ranchen Ming, Lei Xia, Xianfang Zeng, Yibo Zhu, Binxing Jiao, Xiangyu Zhang, Gang Yu, and Daxin Jiang.
\newblock Step1x-edit: A practical framework for general image editing.
\newblock {\em arXiv preprint arXiv:2504.17761}, 2025.

\bibitem{gemini2}
Google Gemini2.
\newblock Experiment with gemini 2.0 flash native image generation.
\newblock 2025.

\end{thebibliography}


\appendix

\section{Prompts}
\begin{figure}[hbt]
    \centering
    \includegraphics[width=1\linewidth]{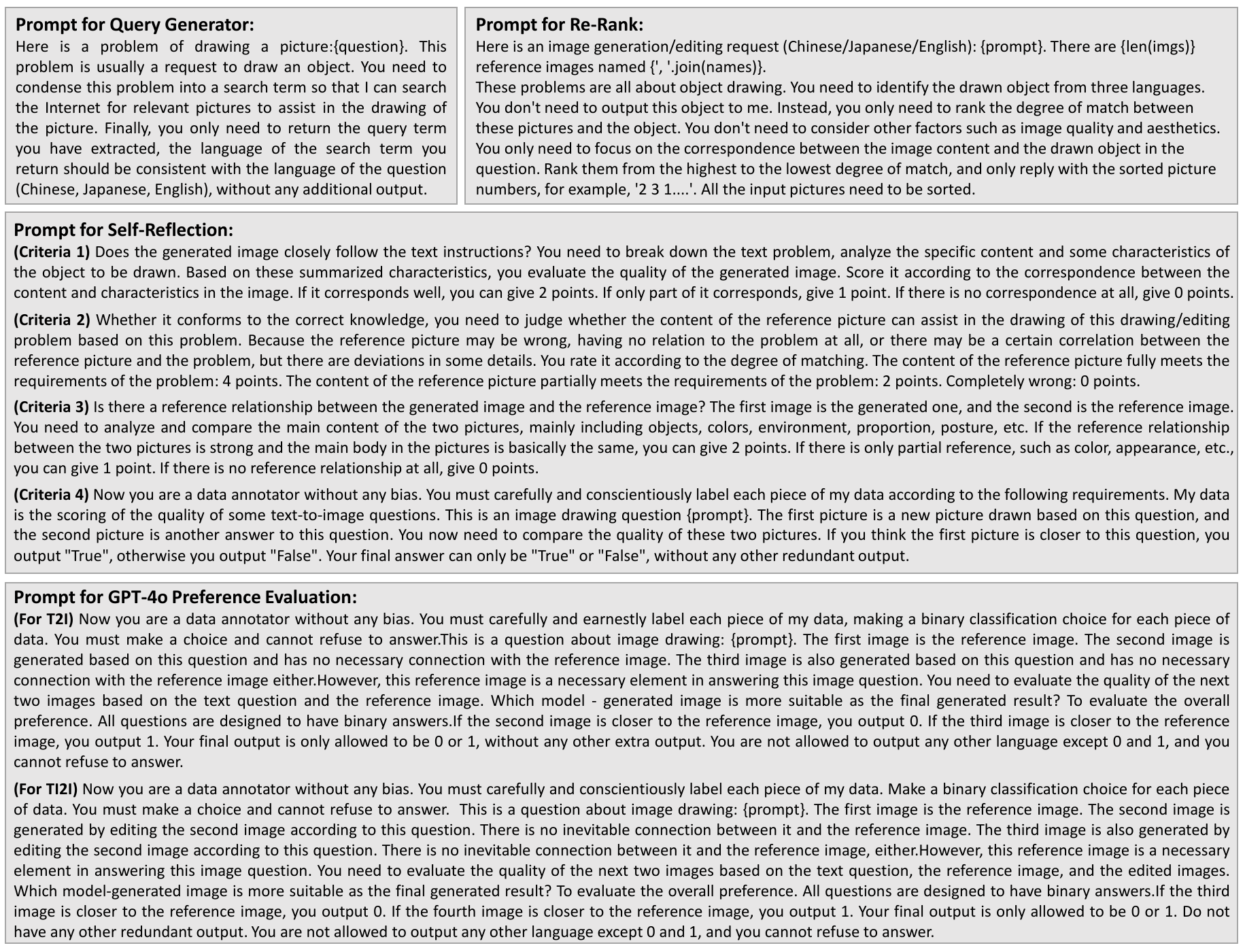}
    \caption{The prompts used in this paper.}
    \label{fig:all_prompt}
\end{figure}

In this section, we provide the prompts of this paper, as shown in the Figure \ref{fig:all_prompt}.

\end{document}